\documentclass[11pt,letterpaper]{article}
\usepackage{emnlp2017}
\usepackage{times}
\usepackage{latexsym}
\usepackage{hyperref}
\usepackage{graphicx}
\usepackage{caption}
\usepackage{url}
\usepackage{mathtools}
\usepackage{amsmath, bm}
\usepackage{resizegather}
\usepackage{subcaption}
\usepackage[english]{babel}
\usepackage[autostyle]{csquotes}
\usepackage{lipsum}

\renewcommand*{\thefootnote}{\fnsymbol{footnote}}

\emnlpfinalcopy

\def\emnlppaperid{***}

\title{Distance-based Self-Attention Network for Natural Language Inference}

\author{Jinbae Im \and Sungzoon Cho \\
        Department of Industrial Engineering \\
        Seoul National University \\
        Seoul, South Korea \\
  {\tt jinbae@dm.snu.ac.kr}, {\tt zoon@snu.ac.kr}}

\date{}

\begin{document}

\maketitle

\begin{abstract}
Attention mechanism has been used as an ancillary means to help RNN or CNN. However, the Transformer~\citep{AttentionAYN} recently recorded the state-of-the-art performance in machine translation with a dramatic reduction in training time by solely using attention. Motivated by the Transformer, Directional Self Attention Network~\citep{DiSAN}, a fully attention-based sentence encoder, was proposed. It showed good performance with various data by using forward and backward directional information in a sentence. But in their study, not considered at all was the distance between words, an important feature when learning the local dependency to help understand the context of input text. We propose Distance-based Self-Attention Network, which considers the word distance by using a simple distance mask in order to model the local dependency without losing the ability of modeling global dependency which attention has inherent. Our model shows good performance with NLI data, and it records the new state-of-the-art result with SNLI data. Additionally, we show that our model has a strength in long sentences or documents.
\end{abstract}

\section{Introduction}

Sequence modeling has been employing Recurrent Neural Networks (RNN) or Convolutional Neural Networks (CNN) mostly. More recently, models incorporating attention mechanisms have shown good performance in machine translation~\citep{NMT1,NMT2}, Natural Language Inference (NLI)~\citep{NLI_INNER}, and Question Answering (QA)~\citep{QA_1,QA_2} etc. Attention mechanisms used to be exploited in conjunction with RNN or CNN as an ancillary means to help improve performance. Lately, \citet{AttentionAYN} presented the first fully attention-based model, which recorded the state-of-the-art result in machine translation. As a fully attention-based model can consider all words in a sentence at once, parallelization leads to great reduction in training time.

Motivated by \citet{AttentionAYN}, \citet{DiSAN} proposed the first fully attention-based sentence encoder. \citet{DiSAN} recorded good performance in a variety of tasks. In particular, they recorded the state-of-the-art result with Stanford Natural Language Inference (SNLI) dataset~\citep{SNLI} which is a representative dataset of NLI. The NLI task aims to classify the relationship between two sentences as entailment, contradiction, or neutral. One of the approaches to solving the NLI task is to use sentence-encoding based models.\footnote[1]{The NLI task can be solved through two different approaches: sentence encoding-based models and joint models. The former separately encode each sentence, whereas the latter take into account the direct relationship between two sentences. Between them, sentence-encoding based models focus on training sentence encoder that can represent sentences in vector form well. We focus on the former approach, since the objective of our work is to develop an advanced sentence-encoding model.} \citet{DiSAN} presented a sentence-encoding based model reflecting directional information in a sentence. However, the distance between words was not considered at all in their model, and the directional information simply involved words before and after the reference word. Altogether, positional information of words was not fully taken into account. As a result, the difference of importance between the distant words and the nearby words was not appropriately reflected. Hence local dependency was not properly modeled, which in turn failed to capture the context information in long sentences.

To tackle this limitation, we propose Distance-based Self-Attention Network which introduces a distance mask which models the relative distance between words. In conjunction with a directional mask, the distance mask allows us to incorporate complete positional information of words in our model. Our Distance-based Self-Attention Network achieved good performance with NLI data, and recorded the state-of-the-art result with SNLI. Our model worked exceptionally well with long sentences, in particular. We also visualized the effect of the distance mask to show that our model can grasp both local dependency and global dependency. 


\section{Related Works}
\label{2}

NLI tasks have been studied through models of various structures. Most of all, models combining attention with Long Short-Term Memory (LSTM) have performed well. \citet{NLI_INNER} improved the performance by adding the mean pooling vector to the conventional attention model in which attention is applied to hidden states of LSTM. \citet{NLI_GATE} used the input gates of the LSTM as attention weights to simplify the model structure. In \citet{NLI_GATE} and \citet{MultiNLI_STACK}, short-cut connections in stacked LSTM, in combination with max-pooling originally suggested by \citet{Facebook}, were proven effective in improving performance, recording the state-of-the-art performance in MultiNLI. And \citet{NLI_NSE} used the memory for sentence encoding motivated by Neural Turing Machine~\citep{NTM}.

\citet{AttentionAYN} was the first study to construct an end-to-end model with attention alone, and recorded the state-of-the-art performance in machine translation tasks. \citet{AttentionAYN}'s encoder-decoder framework consists of a multi-head attention and a position-wise feed forward network as a basic building block which is deeply stacked combined with residual connection. The multi-head attention projects the input sentences to multiple subspaces and then computes the scaled dot-product attention in each subspace. The results in each subspace are then concatenated and projected again. Position-wise feed forward network adds non-linearity to vector representations of each position. In this way, the fully attention-based model was constructed without using RNN or CNN, and the training cost was greatly reduced. 
 
\citet{DiSAN}, a very recent work, constructed a fully attention-based sentence encoder motivated by \citet{AttentionAYN}. They proposed a multi-dimensional attention mechanism that computes the attention by each dimension through modification of additive attention. In addition, their model exploits directional attention as well as fusion gate motivated by bi-directional LSTM. Directional information was reflected by introducing a simple directional mask. By adding a directional mask to the logit of attention, words in a specific direction in the sentence were masked to avoid attention. The extent to which attention results are ultimately reflected was determined through fusion gate. In our study, we construct our model based on \citet{AttentionAYN}'s basic building block, as well as \citet{DiSAN}'s key model structures. In order to model the distance between words, which was not considered in their works, we transform the multi-head attention in \citet{AttentionAYN}, in particular, to fit our objective. Details can be found in section \ref{4}.

\section{Background}
\label{3}

In \citet{AttentionAYN}, the attention function is defined as follows by introducing the concept of query, key, and value. \enquote{An attention function can be described as mapping a query and a set of key-value pairs to an output, where the query, keys, values, and output are all vectors. The output is computed as a weighted sum of the values, where the weight assigned to each value is computed by a compatibility function of the query with the corresponding key~\citep{AttentionAYN}.} The two most commonly used attentions are additive attention~\citep{NMT1, Additive} and dot-product attention~\citep{Dot, QA_2, AttentionAYN}.

\subsection{Additive Attention}

Let query, $i$th key, and $i$th value be $q$, $k_i$, and $v_i$ respectively. ($q\in R^{d_k}$, $k_i\in R^{d_k}$, and $v_i\in R^{d_v}$)

Compatibility function of the query with the $i$th key is represented by the following equation 
\ref{additive_l}.

\begin{equation}
    f(q, k_i) = l_i = u^T\sigma(q+k_i),
    \label{additive_l}
\end{equation}
where $u\in R^{d_k}$, and $\sigma (\cdot)$ is an activation function usually chosen as tanh.

And attention weight assigned to each $i$th value is computed by applying the softmax function to $l_i$ and final output is weighted sum of value as following equations.

\begin{equation}
    {w_i} = {\frac{\exp(l_i)}{\sum_{j=1} \exp(l_j)}}
    \label{additive_2}
\end{equation}
\begin{equation}
    \text{Output} = \sum_{i=1} {w_i}{v_i}
    \label{additive_3}
\end{equation}

\subsection{Dot-product Attention}
Dot-product attention is the same as additive attention except for compatibility function. In dot-product attention, compatibility function is computed by the following equation \ref{dot} in place of the equation \ref{additive_l}.

\begin{equation}
    f(q, k_i) = l_i = \langle\ q, k_i \rangle
    \label{dot}
\end{equation}

On implementation, dot-product attention is much faster and more space-efficient than additive attention due to optimized matrix multiplication.

In practice, however, additive attention outperforms dot product attention for large values of $d_k$. So \citet{AttentionAYN} used scaled dot-product attention instead of normal dot-product attention to prevent performance loss in large dimension as following equation \ref{scaled_dot}.

\begin{equation}
    f(q, k_i) = l_i = \frac{\langle\ q, k_i \rangle}{\sqrt{d_k}}
    \label{scaled_dot}
\end{equation}

\section{Proposed Model}
\label{4}

\subsection{Overall Architecture}

\begin{figure}[h]
\begin{center}
	\includegraphics[width=0.4\textwidth]{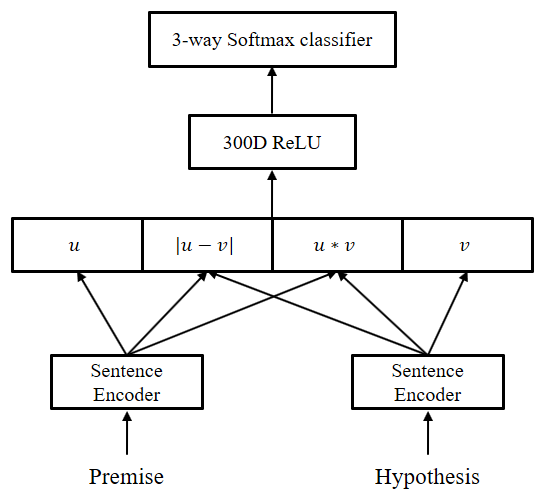}
\caption{\textbf{Overall architecture}}
\label{Overall architecture}
\end{center}
\end{figure}

Our model's overall architecture is shown in Figure \ref{Overall architecture}. We follow the conventional architecture for training NLI data. First, the two input sentences, premise and hypothesis, are encoded as vectors, $u$ and $v$ respectively, through identical sentence encoders. For the encoded vectors $u$ and $v$, the representation of relation between the two vectors is generated by the concatenation of $u$, $v$, $|u-v|$, and $u*v$. Thereafter, a probability for each of the 3-class is generated through the 300D ReLU layer and the 3-way softmax output layer. We configured the model with the setting of 1layer 300D as in \citet{DiSAN} to focus on the performance evaluation of the sentence encoder itself. Layer normalization~\citep{LN} and dropout are applied to 300D ReLU layer.

\subsection{Sentence Encoder}

\begin{figure}[h]
\begin{center}
	\includegraphics[width=0.45\textwidth]{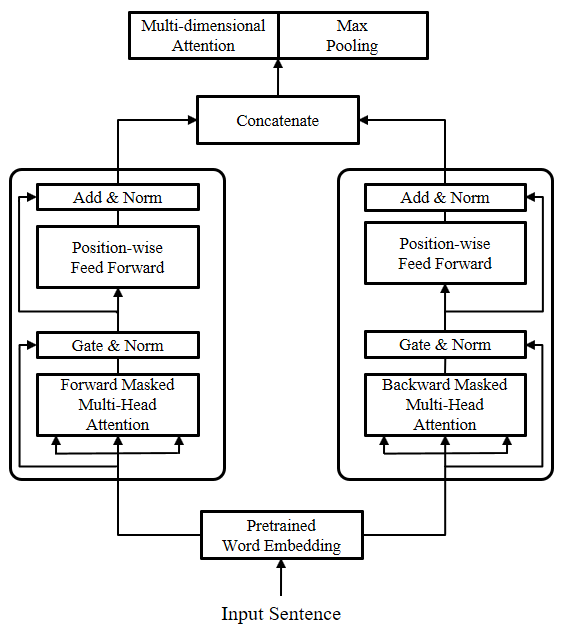}
    
\caption{\textbf{Sentence encoder}}
\label{Sentence encoder}
\end{center}
\end{figure}

The sentence encoder structure proposed in this paper is shown in Figure \ref{Sentence encoder}. The term \enquote{Norm} in Figure \ref{Sentence encoder} stands for layer normalization. The sentence encoder of Figure \ref{Sentence encoder} encodes the premise and hypothesis in a vector form. We describe each component of our sentence encoder in detail in the following subsections.

\subsubsection{Word Embedding Layer}

Let an input sentence be a sequence of discrete words $\mathbf{x} = [x_1, x_2, \cdot\cdot\cdot, x_n]$, where $x_i \in R^N$ is a one-hot representation of the word $i$, and $N$ is the vocabulary size. These one-hot representations are transformed into dense representations by using the pre-trained word embedding.

Let $W_e \in R^{d_e\times N}$ be a pre-trained word embedding matrix. Then a sequence of dense word representations can be written as $\mathbf{w}$ = ${W_e}\mathbf{x}=[w_1, w_2, \cdot\cdot\cdot, w_n]$, where $w_i \in R^{d_e}$ is dense representation of the word $i$.

\subsubsection{Masked Multi-Head Attention}

The masked multi-head attention is a variation of the multi-head attention employed by \citet{AttentionAYN}. The scaled dot-product attention of \citet{AttentionAYN} is expressed as following:

\begin{equation}
    \text{Attention}(Q, K, V) = \text{softmax}(\frac{QK^T}{\sqrt{d_k}})V
    \label{dot-product attention}
\end{equation}
where $Q, K, V$ are matrices composed of a set of queries, keys, and values, respectively.

We transform equation \ref{dot-product attention} and express the masked attention as following:

\begin{equation}
   \begin{aligned}
   \text{Mas}& \text{ked}(Q, K, V)\\
   &= \text{softmax}(\frac{QK^T}{\sqrt{d_k}} + M_{dir} + \alpha M_{dis})V
   \end{aligned}
   \label{masked_attention}
\end{equation}

Here, $M_{dir} \in R^{n\times n}$ is the directional mask as proposed in \citet{DiSAN}, while $M_{dis} \in R^{n \times n}$ is the distance mask proposed in this model. Hyper parameter $\alpha$ is the distance-alpha tuned through validation data.

\begin{figure}[h]
\centering
\begin{subfigure}{.2\textwidth}
  \centering
  \includegraphics[width=\linewidth]{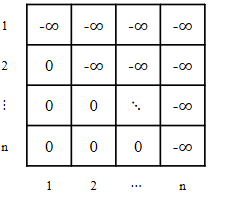}
  \caption{Forward mask}
  \label{Forward_mask}
\end{subfigure}\hspace{.5em}
\begin{subfigure}{.2\textwidth}
  \centering
  \includegraphics[width=\linewidth]{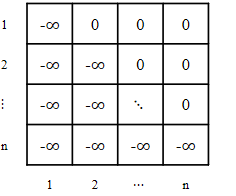}
  \caption{Backward mask}
  \label{Backward_mask}
\end{subfigure}
\caption{\textbf{Directional mask}}
\label{Directional_mask}
\end{figure}

$M_{dir}$ consists of the forward mask and backward mask as explained in Figure \ref{Directional_mask}. In the Forward Masked Multi-Head Attention phase, the forward mask is selected, and in the Backward Masked Multi-Head Attention phase, the backward mask. The forward masks prevent words that appear after a given word from being considered in the attention process, while backward masks prevent words that appear before from consideration by adding $-\infty$ to the logits before taking the softmax at the attention phase. The diagonal component of $M_{dir}$ is also set to $-\infty$ so that each token does not consider itself to attention, and the information of each token is later transmitted through the fusion gate of section ~\ref{fusion_gate}

\begin{figure}[h]
\begin{center}
	\includegraphics[width=0.2\textwidth]{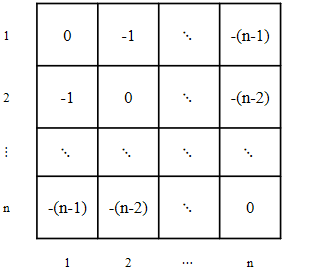}    
\caption{\textbf{Distance mask}}
\label{Distance_mask}
\end{center}
\end{figure}

$M_{dis}$ is shown in the Figure ~\ref{Distance_mask}. The $(i, j)$ component of the distance mask is $-|i-j|$, representing the distance between $(i+1)$th word and $(j+1)$th word multiplied by $-1$. By multiplying this value by $\alpha$ and adding it to logit, the attention weight becomes smaller as distance increases. That is, the distance mask serves to concentrate on the local words around the reference word. Such a structure may appear similar to a CNN filter extracting a local feature. Yet, the big difference is that CNN only uses information in the window size, whereas our model considers all words in a sentence at once, concentrating on the local words by taking account of the relative distance between words. 

By using the distance mask, the distance between words, not considered through the directional mask of \citet{DiSAN}, was considered additionally, so the complete positional information of words was taken into consideration.\footnote[1]{In \citet{AttentionAYN}, the positional information of the word was used through positional encoding. By adding the positional encoding vector to the word embedding vector, the embedding changed according to the absolute position of the word in the sentence. However, in sentence modeling, the relative position with respect to the other words is important, not the absolute position of the word. In other words, what words are placed in order before and after the word is important, not the absolute position of the word in a sentence. Therefore, we take the relative position directly into account in our model through the distance mask instead of the positional encoding which considers the relative position indirectly.}

The masked multi-head attention can be expressed as following:

\begin{equation}
    \begin{aligned}
    \text{Masked}&\text{\_Multi-Head}(Q, K, V) \\
    &= \text{concat}(head_1, \cdots, head_h)W^O
    \end{aligned}
    \label{Masked Multi-head Attention1}
\end{equation}
where $head_i = \text{Masked}(QW_i^Q, KW_i^K, VW_i^V)$, with $h$ as the number of heads, $W_i^Q$, $W_i^K$, $W_i^V$ $\in R^{d_e \times d_e / h}$, and $W^O$ $\in R^{d_e \times d_e}$. $Q, K, V$ $\in R^{n \times d_e}$ are matrices created from $n$ word embedding vectors of sentences and expressed as equation \ref{QKV}.

\begin{equation}
Q=K=V=
\begin{bmatrix} 
- & w_1 & - \\
- & w_2 & - \\
  & \vdots & \\
- & w_n & - 
\end{bmatrix}
\quad
\label{QKV}
\end{equation}

The masked multi-head attention first projects $Q, K, V$ into $h$ subspaces, respectively, and performs masked attention of equation \ref{masked_attention} for each $Q, K, V$ projection combination. The $h$ attention result is concatenated before projection.\footnote[1]{Multi-head attention~\citep{AttentionAYN} is fast and efficient because it is based on dot-product attention. However, multi-dimensional attention~\citep{DiSAN} has a disadvantage in that it consumes a lot of gpu memory because it requires several 4-dimensional tensors on implementation. So, in our model, the multi-head attention was used as a base structure instead of the multi-dimensional attention. In addition, the performance of the actual implementation was also better with multi-head attention.}

\renewcommand*{\thefootnote}{\arabic{footnote}}

\subsubsection{Fusion Gate}
\label{fusion_gate}

At the fusion gate, raw word embedding $S \in R^{n \times d_e}$ and the result of masked multi-head attention $H \in R^{n \times d_e}$ in equation \ref{SH} are used as input. 

\begin{equation}
S = 
\begin{bmatrix} 
- & w_1 & - \\
- & w_2 & - \\
  & \vdots & \\
- & w_n & - 
\end{bmatrix}
H = 
\begin{bmatrix} 
- & h_1 & - \\
- & h_2 & - \\
  & \vdots & \\
- & h_n & - 
\end{bmatrix}
\quad
\label{SH}
\end{equation}

First, we generate $S^F, H^F$ by projecting $S, H$ using $W^S, W^H \in R^{d_e \times d_e}$. Mathematically: 

\begin{equation}
  \begin{gathered}
  S^F = SW^S \\
  H^F = HW^H
  \end{gathered}
\label{SH_proj}
\end{equation}

Then create gate $F$ as shown in equation \ref{Fusion_gate} where $b^F \in R^{d_e}$. 

\begin{equation}
  \begin{aligned}
  \text{Gate}(S,H) &= F \odot{} S^F + (1-F) \odot{} H^F \\
  \text{where} \ F &= \text{sigmoid}(S^F+H^F+b^F)
  \end{aligned}
\label{Fusion_gate}
\end{equation}

 Finally, we obtain the gated sum by using $F$. It is common in many papers including \citet{DiSAN} to use raw $S$ and $H$ in gated sum. We, however, use the gated sum of $S^F$ and $H^F$ which resulted in a significant increase in accuracy.

\subsubsection{Position-wise Feed Forward Networks}

We used position-wise feed forward network structure of \citet{AttentionAYN} as it is. The position-wise feed forward network employs the same fully connected network to each position of sentence, in which the fully connected layer consists of two linear transformations, with the ReLU activation in between. Mathematically:

\begin{equation}
  \text{FFN}(x) = max(0, x W_1^{P} + b_1^{P})W_2^{P} + b_2^{P}
\label{POS_FFN}
\end{equation}
where $x \in R^{1 \times d_e}$, $W_1^{P} \in R^{d_e \times d_{ff}}$, $W_2^{P} \in R^{d_{ff} \times d_e}$, $b_1^{P} \in R^{d_{ff}}$, and $b_2^{P} \in R^{d_e}$.

The FFN function of the above equation \ref{POS_FFN} is applied to each position of the result of the fusion gate. Note that position-wise feed forward network is combined with the residual connection as shown in Figure \ref{Sentence encoder}. That is, FFN learns the residuals. In our model, $d_{ff}$ was set to $4d_e$.

\subsubsection{Pooling Layer}

The vector representation of input sentence is generated through the pooling layer after the concatenation of the results of forward directional self attention and backward directional self attention. That is, the input of pooling layer is $U = [U^{fw};U^{bw}] \in R^{n \times 2d_e}$ where each directional self attention output is $U^{fw} \in R^{n \times d_e}$, $U^{bw} \in R^{n \times d_e}$. 

We use the multi-dimensional source2token self-attention of \citet{DiSAN} for our multi-dimensional self-attention.

For $i$th row vector of $U$, $u_i$, logit $l(u_i)$ is computed as following:

\begin{equation}
  l(u_i) = ELU(u_i W_1^{M} + b_1^{M})W_2^{M} + b_2^{M}
\label{Multi-dim1}
\end{equation}
where $u_i = U_{i*} \in R^{1 \times 2d_e}$, $W_1^{M}, W_2^{M} \in R^{2d_e \times 2d_e}$, and $b_1^{M}, b_2^{M} \in R^{2d_e}$.

The calculations of logit consist of two linear transformations, with the Exponential Linear Units (ELU) activation function~\citep{ELU} in between. Multi-dimensional attention differs from general attention in that the logit for an input vector is not a scalar but a vector with dimensions equal to the dimensions of the input vector. This allows each dimension of the input vector to have a scalar logit, and we can perform attention to $ n $ word tokens in each dimension, as illustrated below by equation \ref{Multi-dim2}, \ref{Multi-dim3}. Note that softmax is performed on the row dimension of $L$, not the column dimension.

\begin{equation}
  \begin{gathered}
  M = \text{softmax}(L) \odot{} U \\
  \text{where} \ L = 
  \begin{bmatrix} 
  - & l(u_1) & - \\
  - & l(u_2) & - \\
    & \vdots & \\
  - & l(u_n) & - 
  \end{bmatrix}
  \quad
  \end{gathered}
\label{Multi-dim2}
\end{equation}

\begin{equation}
  \text{Multi-dimensional}(U) = \sum\limits_{i=1}^n M_{i*}
\label{Multi-dim3}
\end{equation}

The $2d_e$-dimensional output vector of multi-dimensional attention and the $2d_e$-dimensional vector obtained by applying max pooling to $U$ are concatenated to encode the input sentence as a $4d_e$-dimensional vector.

\section{Experiments and Results}

\subsection{Dataset}

The dataset used in the experiments are SNLI ~\citep{SNLI} and MultiNLI ~\citep{MultiNLI} datasets. The SNLI dataset consists of 549,367 / 9,842 / 9,824 (train / valid / test) premise and hypothesis pairs; and the MultiNLI dataset, 392,702 / 9,815 / 9,832 / 9,796 / 9,847 (train / valid\_matched / valid\_mismatched / test\_matched / test\_mismatched) sentence pairs. The two datasets have the same format, but sentences in the MultiNLI dataset are much longer than those in SNLI dataset. In addition, MultiNLI dataset consists of various genre information. If genres included in the train data are also found in valid (test) data, then the dataset is called \enquote{matched}; if valid (test) data includes genres that are not in the train data, then the dataset is called \enquote{mismatched}.

\subsection{Training Details}

We used the Glove 840B 300D\footnote{\tt https://nlp.stanford.edu/projects/glove/} ($d_e = 300$) for the pre-trained word embedding without any fine-tuning. This is to train the more universally usable sentence encoder. 

Layer normalization~\citep{LN} was applied to all linear projections of masked multi-head attention, fusion gate, and multi-dimensional attention. We applied residual dropout as used in \citet{AttentionAYN}, with dropout to the output of masked multi-head attention and $S^F + H^F + b^F$ of fusion gate.

We set $h$ $=$ $5$, $~\alpha$ $=$ $1.5$ in the masked multi-head attention, and the dropout probability was set to 0.1. Batch size was 64, and the model was trained with Adam~\citep{Adam} optimizer, with a learning rate of 0.001. All models were implemented via Tensorflow on single Nvidia Geforce GTX 1080Ti GPU.

\subsection{SNLI Results}

\begin{table*}[t]
\centering
\resizebox{\textwidth}{!}{\begin{tabular}{lcccc}
\hline \multicolumn{1}{c}{\bf Model Name} & $ \bm{|~\theta|}$ & \bf T(s)/epoch & \bf Train Acc(\%) & \bf Test Acc(\%) \\ \hline
\bf Feature-based models \\ \hline
Unlexicalized features~\citep{SNLI} & & & 49.4 & 50.4 \\
+Unigram and bigram features~\citep{SNLI} & & & 99.7 & 78.2 \\
\hline
\bf Sentence encoding-based models \\ \hline
100D LSTM encoders~\citep{SNLI}& 220k & & 84.8 & 77.6 \\
300D LSTM encoders~\citep{NLI_SPIN} & 3.0m & & 83.9 & 80.6 \\
1024D GRU encoders~\citep{NLI_VENDROV} & 15m & & 98.8 & 81.4 \\
300D Tree-based CNN encoders~\citep{NLI_TREE} & 3.5m & & 83.3 & 82.1 \\
300D SPINN-PI encoders~\citep{NLI_SPIN} & 3.7m & & 89.2 & 83.2 \\
600D Bi-LSTM encoders~\citep{NLI_INNER} & 2.0m & & 86.4 & 83.3 \\
300D NTI-SLSTM-LSTM encoders~\citep{NLI_NTI} & 4.0m & & 82.5 & 83.4 \\
600D Bi-LSTM encoders+intra-attention~\citep{NLI_INNER} & 2.8m & & 84.5 & 84.2 \\
300D NSE encoders~\citep{NLI_NSE} & 3.0m & &86.2 & 84.6 \\
600D Deep Gated Attn. BiLSTM encoders~\citep{NLI_GATE} & 11.6m & & 90.5 & 85.5 \\
600D Directional Self-Attention Network~\citep{DiSAN} & 2.4m & 587 & 91.1 & 85.6 \\ \hline
Our self-attention network (without distance mask) & 4.7m & 687 & 88.1 & 86.0 \\
Our Distance-based Self-Attention Network & 4.7m & 693 & 89.6 & \bf{86.3} \\ \hline
\end{tabular}}
\caption{\label{SNLI_TABLE} \textbf{Experimental results of different models on SNLI data.} $|~\theta|$ : number of parameters (excluding word embedding part). T(s)/epoch : average training time (second) per epoch.}
\end{table*}

Experimental results of SNLI data compared with the existing models on the SNLI leader-board\footnote{\tt https://nlp.stanford.edu/projects/snli/} are shown in Table \ref{SNLI_TABLE}. Compared with the existing state-of-the-art model~\citep{DiSAN}, the number of parameters and the training time increased, but our results show the new state-of-the-art record. We also looked at the model with distance mask removed to verify the effect of the distance mask proposed in this paper. Results show that the addition of the distance mask improved the performance without significantly affecting the training time or increasing the number of parameters. 

\begin{figure}[!h]
\begin{center}
	\includegraphics[width=0.35\textwidth]{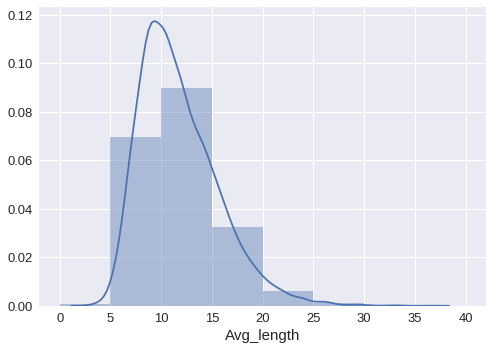}    
\caption{\textbf{SNLI average sentence length}}
\label{snli_length}
\end{center}
\end{figure}

\begin{figure}[!h]
\begin{center}
	\includegraphics[width=0.45\textwidth]{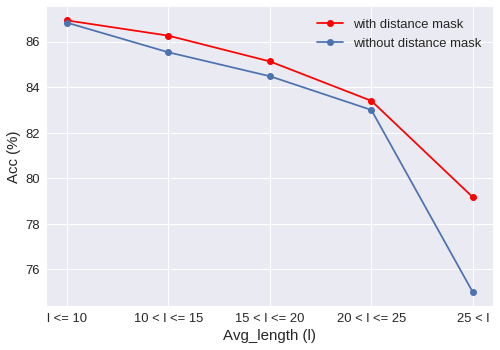}    
\caption{\textbf{With distance mask vs. Without distance mask.} Change of test accuracy on SNLI data w.r.t average length of sentence pair.}
\label{snli_length_graph}
\end{center}
\end{figure}

The improvement of the test accuracy by introducing the distance mask is only by 0.3\% point, potentially because SNLI data mostly consist of short sentences. Hence, we additionally examined how the effect of the distance mask changes as the average length of the two sentences of premise and hypothesis pair changes. The distribution of the average length of the two sentences of the SNLI test data is shown in Figure \ref{snli_length}, and the effect of the distance mask according to the average length change can be seen from Figure \ref{snli_length_graph}. Figure \ref{snli_length_graph} shows that the accuracy is similar until the average length is less than 25, yet the test accuracy of the model without the distance mask deteriorates drastically for data of an average length exceeding 25. This demonstrates that the distance mask has an advantage with long sentences or documents.

\subsection{MultiNLI Results}

\begin{table*}[t]
\centering
\resizebox{\textwidth}{!}{\begin{tabular}{lcccc}
\hline \multicolumn{1}{c}{\bf Model Name}& \bf SNLI Mix & $ \bm{|~\theta|}$ & \bf Matched Test Acc(\%) & \bf Mismatched Test Acc(\%) \\ \hline
\bf Baseline \\ \hline
CBOW~\citep{MultiNLI} & O & & 66.2 & 64.6 \\
BiLSTM~\citep{MultiNLI} & O & & 67.5 & 67.1 \\
\hline
\textbf{RepEval 2017}~\citep{RepEVAL}\\ \hline
Cha-level Intra-attention BiLSTM encoders ~\citep{MultiNLI_CHA}& O & & 67.9 & 68.2 \\
BiLSTM + enhanced embedding + max pooling~\citep{MultiNLI_VU}  & X & & 70.7 & 70.8 \\
BiLSTM + Inner-attention~\citep{MultiNLI_ATTEN} & O & & 72.1 & 72.1 \\
Deep Gated Attn. BiLSTM encoders~\citep{NLI_GATE} & X &11.6m & 73.5 & \bf 73.6 \\
Shortcut-Stacked BiLSTM encoders~\citep{MultiNLI_STACK} & O & 140.2m & \bf 74.5 & 73.5 \\ \hline
\bf Fully attention-based models \\ \hline
Directional Self-Attention Network~\citep{DiSAN} &X & 2.4m & 71.0 & 71.4 \\ \hline
Our Distance-based Self-Attention Network
& X &4.7m& 74.1 & 72.9 \\ \hline
\end{tabular}}
\caption{\label{MULTINLI_TABLE} \textbf{Experimental results of different models on MultiNLI data.} SNLI Mix : use of SNLI training dataset. $|~\theta|$ : number of parameters (excluding word embedding part).}
\end{table*}

The results of applying SNLI best model to MultiNLI dataset without additional parameter tuning are presented in Table \ref{MULTINLI_TABLE}. Note that matched-test accuracy and mismatched-test accuracy were obtained by submitting our
test results to Kaggle open evaluation platforms: MultiNLI
Matched Open Evaluation\footnote{\scriptsize {\tt https://www.kaggle.com/c/\\ \-\hspace{.75cm} multinli-matched-open-evaluation}} and MultiNLI Mismatched Open
Evaluation\footnote{\scriptsize {\tt https://www.kaggle.com/c/ \\ \-\hspace{.75cm} multinli-mismatched-open-evaluation}}. First, the average test accuracy difference is greater than 2\% when compared to the Directional Self-Attention Network~\citep{DiSAN}. This once again confirms our model's advantage in long sentences, given that the sentence is much longer in MultiNLI.

Compared with the result of RepEVAL 2017~\citep{RepEVAL}, we can see that the Distance-based Self-Attention Network performs well. When compared with the model of \citet{NLI_GATE}, our model showed similar average test accuracy with much lower number of parameters. Also, considering that the model of \citet{NLI_GATE} is a complex LSTM model, our model has an advantage in training time as a fully attention-based model.

\citet{MultiNLI_STACK} showed the best performance with 74.5\% accuracy in Matched Test. However, it is a very deep structured LSTM model with 140.2m parameters. In our model, the inference layer is simply composed of 1 layer of 300D in order to focus on the training of sentence encoder. Both in \citet{NLI_GATE} and \citet{MultiNLI_STACK} models, the inference layer was set very complex in order to improve the MultiNLI accuracy. Taking this into consideration, it can be seen that our Distance-based Self-Attention Network performs competitively given its simpler structure.

\subsection{Case Study}

A case study was conducted to investigate the role of each structure of the Distance-based Self-Attention Network. For this, a sentence \enquote{A lady stands outside of a Mexican market.} is picked among the premise sentences of SNLI test data. We focused on training encoders that can represent each sentence in a vector form well. Therefore, a case study was conducted on a single sentence, not a sentence pair.

\textbf{Masked Multi-Head Attention} We first look at the attention weights in masked multi-head attention. Attention weights represent a $n$ by $n$ matrix corresponding to softmax$(\frac{QK^T}{\sqrt{d_k}} + M_{dir} + \alpha M_{dis})$ of equation \ref{masked_attention}, which is different for each head. Here we look at the average attention weights obtained by averaging the attention weights of each head. The attention weights for each head can be found in Appendix.

\begin{figure}[h]
\centering
\begin{subfigure}{.25\textwidth}
  \centering
  \includegraphics[width=\linewidth]{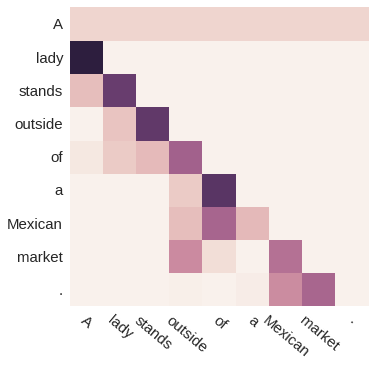}
  \caption{Forward}
  \label{Forward_average_head}
\end{subfigure}
\begin{subfigure}{.25\textwidth}
  \centering
  \includegraphics[width=\linewidth]{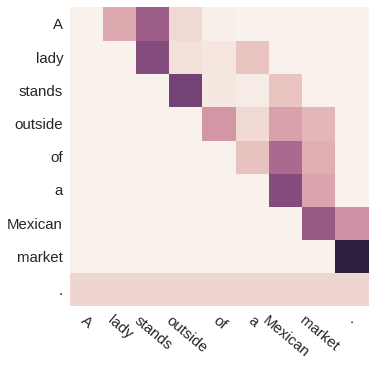}
  \caption{Backward}
  \label{Backward_average_head}
\end{subfigure}
\caption{\textbf{Masked multi-head average attention weights}}
\label{masked multi-head average attention weight}
\end{figure}

The row of the matrix of Figure \ref{masked multi-head average attention weight} represents each word of the sentence, and the column represents the attention weights for each word at each row. It can be seen that the attention weights are heavier to the nearby words as compared to those distant from the reference word. At the same time, \lq outside\rq \ in the forward mask and \lq Mexican\rq \ in the backward mask have high attention weights for several words. From this, it can be seen that important word is considered in the attention process.

\begin{figure}[h]
\centering
\begin{subfigure}{.25\textwidth}
  \centering
  \includegraphics[width=\linewidth]{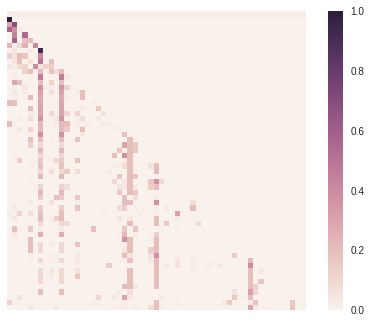}
  \caption{Forward}
  \label{without_forward_mask}
\end{subfigure}
\begin{subfigure}{.25\textwidth}
  \centering
  \includegraphics[width=\linewidth]{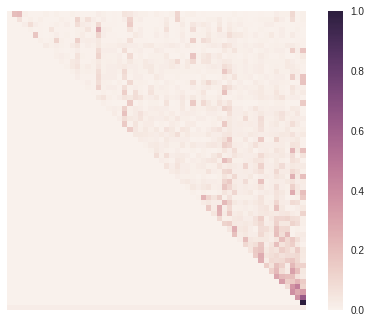}
  \caption{Backward}
  \label{without_backward_mask}
\end{subfigure}
\begin{subfigure}{.25\textwidth}
  \centering
  \includegraphics[width=\linewidth]{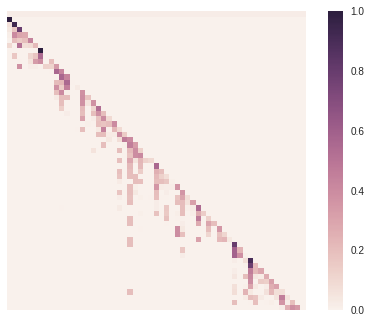}
  \caption{Forward}
  \label{with_forward_mask}
\end{subfigure}
\begin{subfigure}{.25\textwidth}
  \centering
  \includegraphics[width=\linewidth]{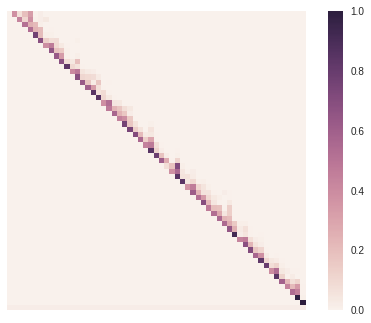}
  \caption{Backward}
  \label{with_backward_mask}
\end{subfigure}
\caption{\textbf{Masked multi-head average attention weights : without/with distance mask.} (a), (b) : without distance mask. (c), (d) : with distance mask}
\label{with/without distance mask}
\end{figure}

\textbf{Distance Mask} We compared the masked multi-head average attention weights for the longest sentence example in the SNLI test data, with length of 57 words to further verify the effect of the distance mask. Panels (a) and (b) of Figure \ref{with/without distance mask} show results without considering distance, while (c) and (d) show the results with the distance mask. In  panels (a) and (b), very distant words are considered in the attention and the overall attention weights were reduced. This implies that each word does not focus on the important words in the attention process, but rather takes into account almost every word, resulting in noisier figures.

However, in panels (c) and (d), the neighboring words are seen more intensively, which implies that the local dependency has been well captured by our model. In addition, as shown in panel (c), even if the word is far apart, it is still considered in the attention process if it is important. This demonstrates the effectiveness of the distance mask to identify local dependencies without losing the ability to grasp the global dependency.

\begin{figure}[h]
\begin{center}
	\includegraphics[width=0.5\textwidth]{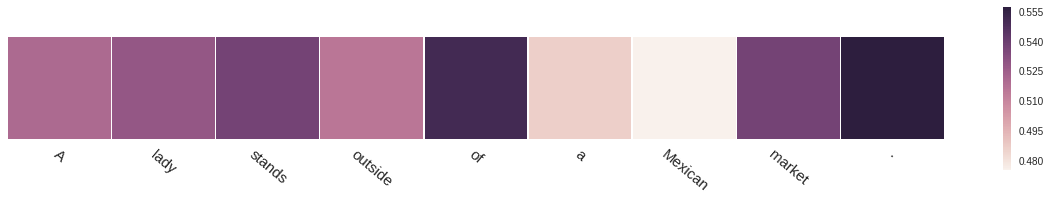}
    
\caption{\textbf{Fusion gate (forward)}}
\label{forward_fusion_gate}
\end{center}
\end{figure}

\textbf{Fusion Gate} We visualize the role of the fusion gate $F \in R^{n \times d_e}$ at forward directional self attention. Figure \ref{forward_fusion_gate} represents the average gate value that averages $d_e$-dimensional gate value for each word. If look at the results of both extremes, keyword \lq Mexican\rq \ has a low gate value, resulting in an output that greatly reflects the multi-head attention result. In contrast, \lq of\rq, \lq .\rq , the words of little importance, have large gate values, which indicates that the original word embedding is greatly reflected, not the multi-head attention result.

\begin{figure}[h]
\centering
\begin{subfigure}{.5\textwidth}
  \centering
  \includegraphics[width=\linewidth]{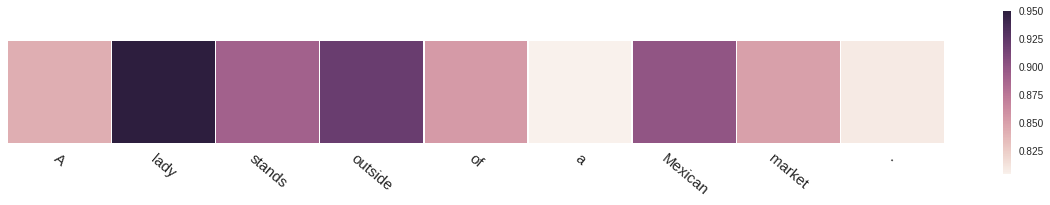}
  \caption{First hidden layer deactivation ratio}
  \label{posffn_1hidden}
\end{subfigure}
\begin{subfigure}{.5\textwidth}
  \centering
  \includegraphics[width=\linewidth]{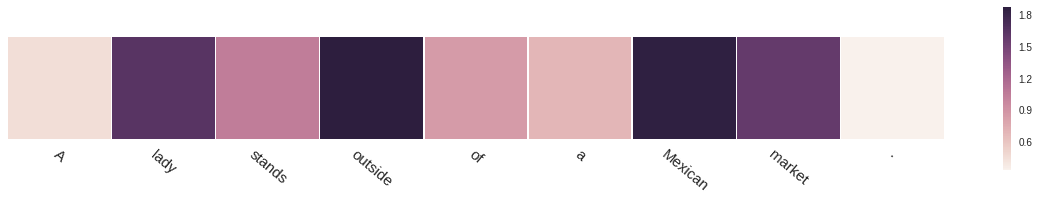}
  \caption{Final output max value (+residual connection)}
  \label{posffn_final}
\end{subfigure}
\caption{\textbf{Position-wise ffn (forward)}}
\label{Posffn_}
\end{figure}

\textbf{Position-wise FFN} For the FFN function of equation \ref{POS_FFN}, Figure \ref{Posffn_}(a) represents the deactivation ratio in the first hidden layer of position-wise ffn.

As shown in Figure \ref{Sentence encoder}, position-wise ffn is used in conjunction with a residual connection. That is, the final output of position-wise ffn for input $x$ is the $d_e$-dimensional vector of LayerNorm$(x+\text{FFN}(x))$. Figure \ref{Posffn_}(b) visualizes the maximum value of this final output vector.

In Figure \ref{Posffn_}, keywords with a high deactivation ratio is shown in panel (a) and a high final max value in panel (b). In case of a word corresponding to a keyword, deactivation occurs frequently in (a), and residual learning is hardly achieved in the position-wise ffn, so that the output of the fusion gate is almost maintained. On the other hand, in case of non-important words, residual learning is performed in position-wise ffn because there is less deactivation in (a), so that the max value of final output becomes smaller in (b). This results in preventing non-important words from consideration in the subsequent pooling layer. In summary, position-wise ffn plays a key role in ensuring that non-critical words are paid less attention to in pooling layers.

\begin{figure}[h]
\centering
\begin{subfigure}{.5\textwidth}
  \centering
  \includegraphics[width=\linewidth]{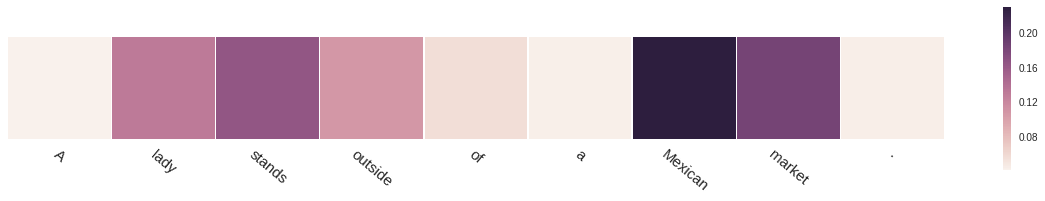}
  \caption{Multi-dimensional attention average weight}
  \label{multidim_attention}
\end{subfigure}
\begin{subfigure}{.5\textwidth}
  \centering
  \includegraphics[width=\linewidth]{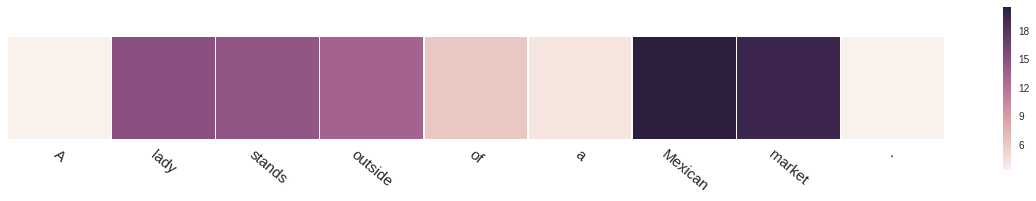}
  \caption{Max pooling ratio (\%)}
  \label{maxpool}
\end{subfigure}
\caption{\textbf{Pooling layer}}
\label{pooling}
\end{figure}

\textbf{Pooling Layer} For the multi-dimensional attention corresponding to Figure \ref{pooling}(a), we visualized the attention weights averaged for each word, where attention weights correspond to softmax$(L) \in R^{n \times 2d_e}$ in equation \ref{Multi-dim2}.

In max pooling, the max value is selected for each column of $U\in R^{n\times 2d_e}$. Thus, in Figure \ref{pooling}(b), we visualize the percentage at which each word is selected in the max pooling operation for the $2d_e$ dimension.

It can be seen that panels (a) and (b) of Figure \ref{pooling} are similar on the whole. In other words, both multi-dimensional attention and max pooling utilize information about key words intensively. A similar result can be expected by using only one of the pooling layers. However, experiment results show that using both multi-dimensional attention and max pooling layer gives better performance.

\section{Conclusion}

In this paper, we propose the Distance-based Self-Attention Network reflecting the distance between words. By reflecting the word distance information, our model learns the local dependency without losing the ability to capture the global dependency. This was achieved through a simple distance mask, so that the performance of the NLI task could be improved while maintaining the number of parameters and training time. In particular, we recorded the new state-of-the-art performance for SNLI data. The introduction of the distance mask improves the performance with longer sentences.

As the research on universal sentence encoders using NLI data was proposed by \citet{Facebook}, we plan to carry out research on fully attention-based networks for universal sentence embedding as future work. We will also study the fully attention-based network in image data and speech data. Especially, regarding image data, capsule network~\citep{Capsule} recently proposed, and as research on new structure to replace CNN is going on, our future work will move in similar directions.


\section*{Acknowledgments}

We would like to thank Hyejin Lee, Hyunjoong Kim, Taewook Kim, Jinwon An, Inbeom Park, Minki Chung, and many others in
SNUDM center, for critical feedback and discussions.

This work was supported by the BK21 Plus Program(Center for Sustainable and Innovative Indus- trial Systems, Department of Industrial Engineering \& Institute for Industrial Systems Innovation, Seoul National University) funded by the Ministry of Education, Korea (No. 21A20130012638), the National Research Foundation (NRF) grant funded by the Korea government (MSIP) (No. 2011- 0030814), and the Institute for Industrial Systems Innovation of SNU.

\bibliography{emnlp2017}
\bibliographystyle{emnlp_natbib}

\newpage

\section*{Appendix}
\label{appendix}

\begin{figure}[h]
\centering
\begin{subfigure}{.25\textwidth}
  \centering
  \includegraphics[width=\linewidth]{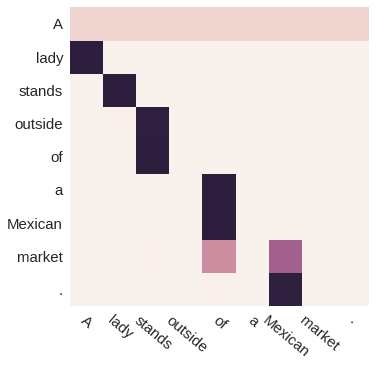}
  \caption{head0}
  \label{forward_head0}
\end{subfigure}
\begin{subfigure}{.25\textwidth}
  \centering
  \includegraphics[width=\linewidth]{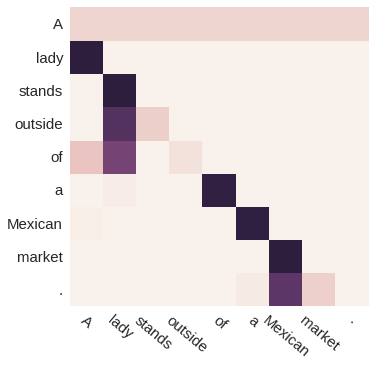}
  \caption{head1}
  \label{forward_head1}
\end{subfigure}
\begin{subfigure}{.25\textwidth}
  \centering
  \includegraphics[width=\linewidth]{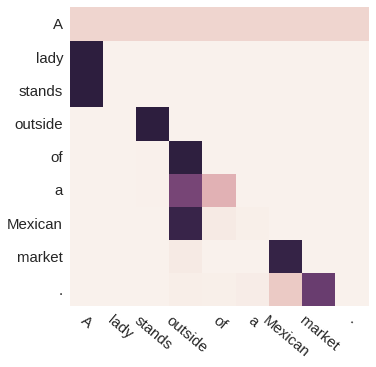}
  \caption{head2}
  \label{forward_head2}
\end{subfigure}
\begin{subfigure}{.25\textwidth}
  \centering
  \includegraphics[width=\linewidth]{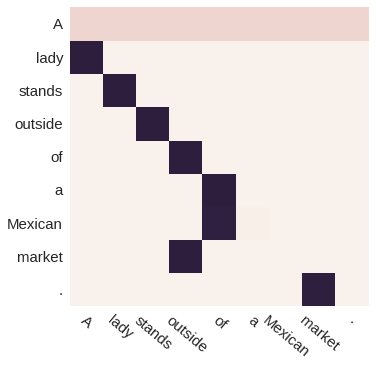}
  \caption{head3}
  \label{forward_head3}
\end{subfigure}
\begin{subfigure}{.25\textwidth}
  \centering
  \includegraphics[width=\linewidth]{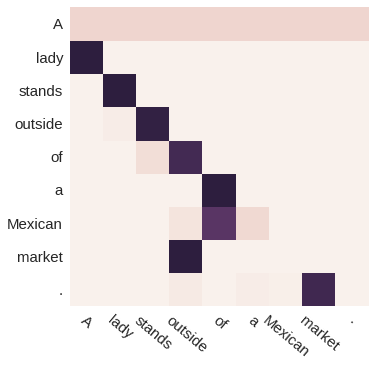}
  \caption{head4}
  \label{forward_head4}
\end{subfigure}
\caption{\textbf{Forward masked multi-head attention weights}}
\label{forward_head}
\end{figure}

\begin{figure}[t!]
\centering
\begin{subfigure}{.25\textwidth}
  \centering
  \includegraphics[width=\linewidth]{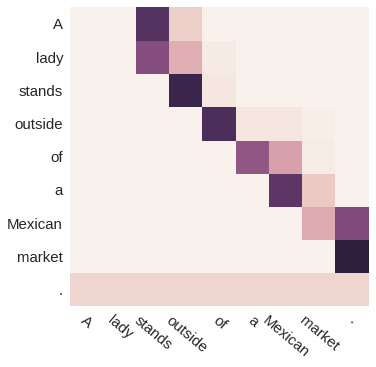}
  \caption{head0}
  \label{backward_head0}
\end{subfigure}
\begin{subfigure}{.25\textwidth}
  \centering
  \includegraphics[width=\linewidth]{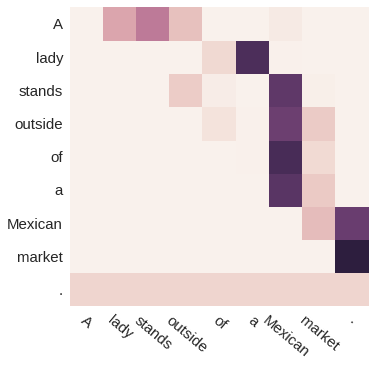}
  \caption{head1}
  \label{backward_head1}
\end{subfigure}
\begin{subfigure}{.25\textwidth}
  \centering
  \includegraphics[width=\linewidth]{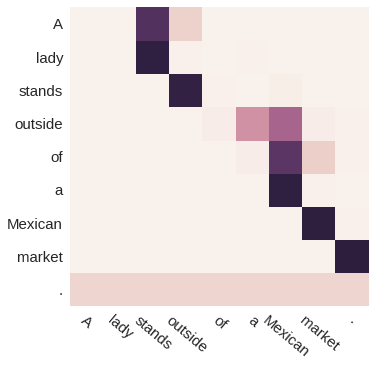}
  \caption{head2}
  \label{backward_head2}
\end{subfigure}
\begin{subfigure}{.25\textwidth}
  \centering
  \includegraphics[width=\linewidth]{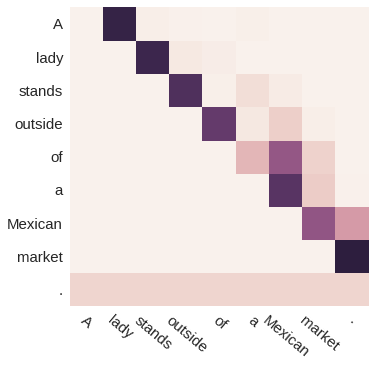}
  \caption{head3}
  \label{backward_head3}
\end{subfigure}
\begin{subfigure}{.25\textwidth}
  \centering
  \includegraphics[width=\linewidth]{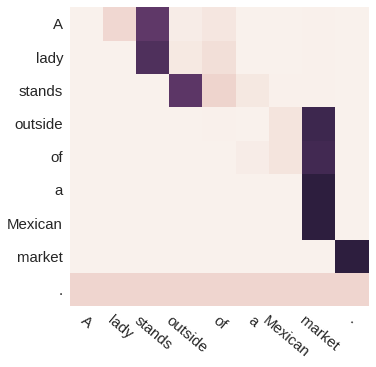}
  \caption{head4}
  \label{backward_head4}
\end{subfigure}
\caption{\textbf{Backward masked multi-head attention weights}}
\label{backward_head}
\end{figure}

\textbf{Masked Multi-Head Attention} The attention weights for each head in the masked multi-head attention are shown in Figures \ref{forward_head} and \ref{backward_head}. Figure \ref{forward_head} shows the result of using a forward directional mask, and Figure \ref{backward_head} is the result of using a backward directional mask. It can be seen that the attention weights are different for each head. This allows our model to capture various dependencies between words in a sentence.

\newpage
\

\end{document}